\pdfoutput=1
\documentclass{article}

\usepackage{PRIMEarxiv}
\usepackage[utf8]{inputenc} 
\usepackage[T1]{fontenc}    
\usepackage{hyperref}       
\usepackage{url}            
\usepackage{booktabs}       
\usepackage{amsfonts}       
\usepackage{nicefrac}       
\usepackage{microtype}      
\usepackage{lipsum}
\usepackage{fancyhdr}       
\usepackage{graphicx}       
\graphicspath{{media/}}     
\usepackage[hang,flushmargin]{footmisc}

\title{The Effects of Data Imbalance Under a Federated Learning Approach for Credit Risk Forecasting}

\author{
  \vspace{0.3em}
  Shuyao Zhang, Jordan Tay, Pedro Baiz \\
  \vspace{0.3em}
  Department of Computing\\
  \vspace{0.3em}
  Imperial College London \\
  \vspace{0.3em}
  London SW7 2AZ, UK 
}

\begin{document}
\maketitle
\let\thefootnote\relax\footnotetext{Authors’ address: Shuyao Zhang, shuyao.zhang18@alumni.imperial.ac.uk; Jordan Tay, duan.tay19@imperial.ac.uk; Pedro Baiz, p.m.baiz@imperial.ac.uk}

\begin{abstract}
Credit risk forecasting plays a crucial role for commercial banks and other financial institutions in granting loans to customers and minimise the potential loss. However, traditional machine learning methods require the sharing of sensitive client information with an external server to build a global model, potentially posing a risk of security threats and privacy leakage. A newly developed privacy-preserving distributed machine learning technique known as Federated Learning (FL) allows the training of a global model without the necessity of accessing private local data directly. This investigation examined the feasibility of federated learning in credit risk assessment and showed the effects of data imbalance on model performance. Two neural network architectures,  Multilayer Perceptron (MLP) and Long Short-Term Memory (LSTM), and one tree ensemble architecture, Extreme Gradient Boosting (XGBoost), were explored across three different datasets under various scenarios involving different numbers of clients and data distribution configurations. We demonstrate that federated models consistently outperform local models on non-dominant clients with smaller datasets. This trend is especially pronounced in highly imbalanced data scenarios, yielding a remarkable average improvement of 17.92\% in model performance. However, for dominant clients (clients with more data), federated models may not exhibit superior performance, suggesting the need for special incentives for this type of clients to encourage their participation.
\end{abstract}

\section{Introduction}

Credit rating is an assessment of the creditworthiness of a prospective debtor, such as an individual, a company or a government. It assists banks to evaluate the ability of a loan applicant to repay its financial obligations and the likelihood of defaulting on these obligations. The traditional approach to predict a borrower's credit risk relies on various subjective and quantitative factors. Such an approach, however, suffers from an unclear contribution of subjective and quantitative analysis to the final credit scoring\cite{smith2003neural}. This limitation can be overcome by machine learning models. A wide range of machine learning models have been extensively employed for credit risk forecasting, such as decision trees\cite{madaan2021loan}, logistic regression\cite{stepanova2001phab}\cite{LAITINEN199997}\cite{steenackers1989credit} and deep neural networks\cite{golbayani2020application}\cite{wang2018deep}. The architecture of these models allows for more accurate and comprehensive quantitative analysis and better prediction results, since they require vast amounts of data that a single organisation may not be able to provide. 

These classic machine learning algorithms, however, also have their weakness. Privacy and confidentiality concerns are the key limitations for most machine learning applications. These algorithms typically require the transmission of local data to the centralised server for model training, and therefore, are susceptible to security threats and data leaks, especially when the training dataset includes sensitive user information, such as financial transactions. In addition, data transfer among various organisations is not always desirable and feasible due to strict regulations and controls on data collection and usage. For example, the General Data Protection Regulation (GDPR) was implemented by the European Union in 2018. Furthermore, moving large volumes of data from local organisations to the central server can result in substantial communication costs. Therefore, there is still a need to develop credit risk forecasting models that are capable of training the model collaboratively while leaving data instances at the providers locally.

Federated Learning (FL) has emerged as a promising solution due to its ability to mitigate privacy and security concerns. It enables the training of models on local clients without sending the sensitive data to a centralised server. We explore the potential benefits and applications of federated learning for credit risk forecasting, with the integration of neural network based models, such as Multilayer Perceptron (MLP) and Long Short-Term Memory (LSTM) , and tree ensemble techniques such as Extreme Gradient Boosting (XGBoost).

We evaluate the effectiveness of federated models relative to local models and investigate the impact of data imbalance on model performance. We particularly focus on both dominant clients with substantial datasets and non-dominant clients with limited datasets, with the aim of measuring model performance improvement after the integration of federated learning. To assess the robustness of federated learning in scalability and non-IID (identically and independently distributed) scenarios, we also compare federated models against centralised models. These experiments include a range of client numbers, from 2 to 10, and consider both balanced and imbalanced data distributions. This study significantly advances both credit risk assessment and federated learning. The key contributions are listed below:

\begin{itemize}

\item To the best of our knowledge, this is the first comprehensive research (see Table~\ref{tab:workcomparison}) to explore the effectiveness of federated learning in credit risk forecasting and investigate the effects of imbalanced data distributions in federated learning. We evaluate the performance by considering a various number of clients and a variable data distribution, and compare the federated models with centralised and local models.

\item This work develops three type of models (MLP, LSTM and XGBoost) within a federated learning framework for credit risk forecasting, which expands the scope beyond the prevailing emphasis on logistic regression models.

\item This work analyses three datasets for both multiclass classification and binary classification while previous credit risk forecasting studies based on federated learning have predominantly focused on one binary classification dataset.

\end{itemize}

\begin{table}
\caption{Comparison of this work and related work in credit risk forecasting under a federated learning approach.}
\label{tab:workcomparison}
\centering
\resizebox{\textwidth}{!}{
\begin{tabular}{@{}ccccccc@{}}
\toprule
Literature &
  \begin{tabular}[c]{@{}c@{}}Number of \\ Datasets\end{tabular} &
  \begin{tabular}[c]{@{}c@{}}Classification\\ Type\end{tabular} &
  \begin{tabular}[c]{@{}c@{}}Number of \\ Clients\end{tabular} &
  \begin{tabular}[c]{@{}c@{}}Number of \\ Model Type\end{tabular} &
  Model Type & \begin{tabular}[c]{@{}c@{}}Imbalanced \\ Data  Distribution\end{tabular}\\ \midrule
  
Zheng et al.\cite{zheng2020vertical} & 1 & Binary & 2    & 1 & Logistic Regression   & No                                                                                   \\
Imateaj and Amini\cite{imteaj2022leveraging} & 1 & Binary & 4, 8 and 12 & 1 & Logistic Regression   &No                                                                                    \\
Li et al.\cite{li2023research} & 1 & Binary & 2    & 1 & Logistic Regression  &No                                                                                     \\
He et al.\cite{he2023privacy} & 1 & Binary & 3    & 3 & \begin{tabular}[c]{@{}c@{}}Logistic Regression \\ Random Forest \\ Extreme Gradient Boosting\end{tabular} & No\\
This work &
  3 &
  \begin{tabular}[c]{@{}c@{}}Binary and\\ multiclass\end{tabular} &
  2, 3, 5 and 10 &
  3 &
  \begin{tabular}[c]{@{}c@{}}Multilayer Perceptron\\ Long Short Term Memory\\ Extreme Gradient Boosting\end{tabular} 
  & Yes \\
  \bottomrule
\end{tabular}
}
\end{table}

\section{Background and Related Work}
\subsection{Overview of Federated Learning}
\subsubsection{Architecture}
\label{model_architecture}
Federated Learning (FL) was first introduced by Google in 2016\cite{mcmahan2017communication}. This novel framework allows multiple clients to collaboratively learn a model on a central server while keeping client data stored locally on their devices. A typical federated learning process is illustrated in Figure~\ref{fig:FL}. The training steps are as follows:

\begin{figure}
\centering
\includegraphics[width = 0.7\hsize]{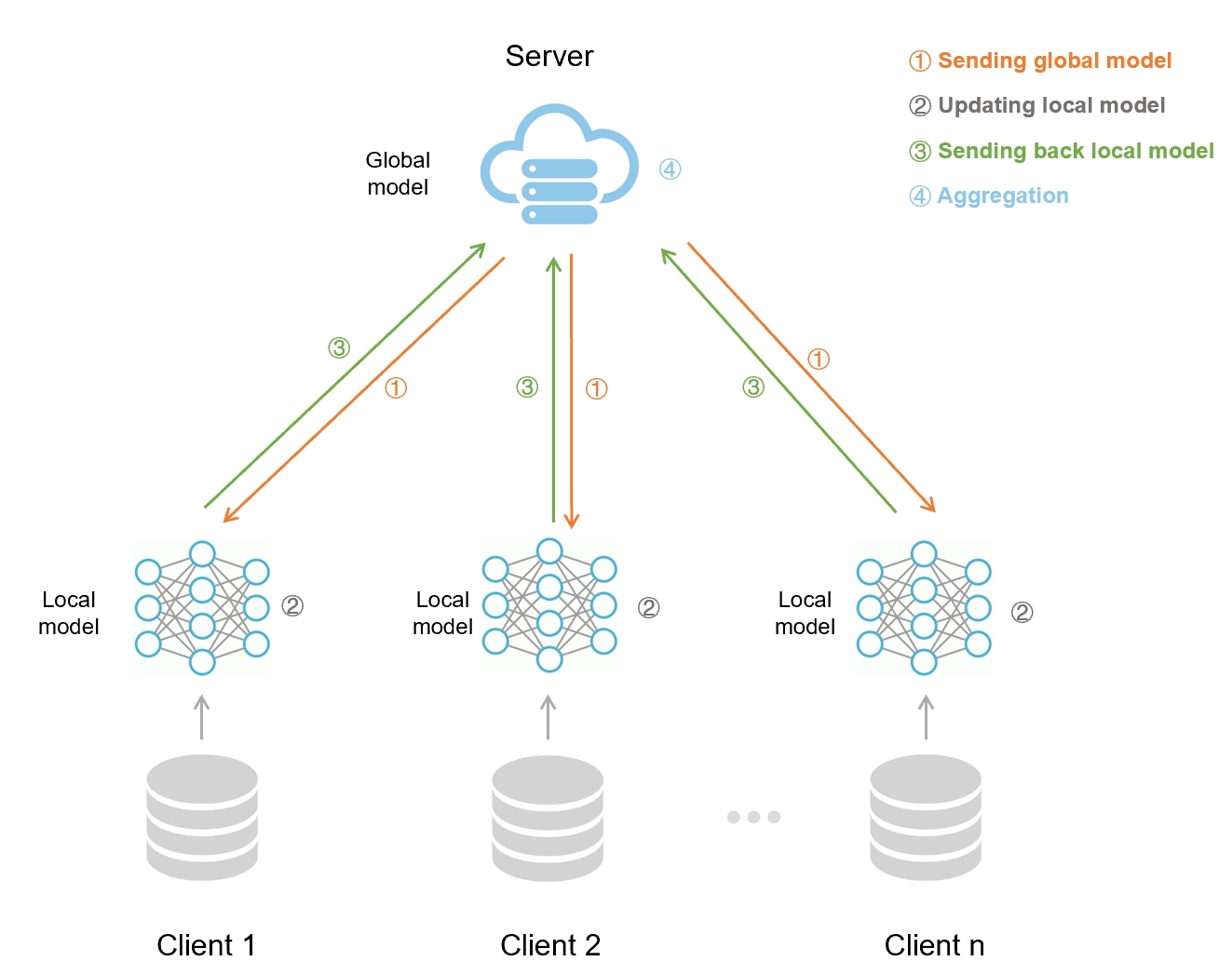}
\caption{Overview of federated learning}
\label{fig:FL}
\end{figure}

1. Broadcast: A global model is first initialised and broadcast from a central server to each participating client.

2. Client computation: Each client updates the model parameters over their local data and learns a personalised model.

3. Aggregation: All local models are sent back to the central server and aggregated by averaging for a joint global model.

4. Iterative model updates: The global model is broadcast to all clients and these steps are repeated until convergence.

The FedAvg algorithm\cite{mcmahan2017communication} is the most widely adopted federated learning algorithm. Its goal is to train a single global model $\theta$ that minimises the empirical risk function over the collective data from all participating clients\cite{kairouz2021advances}, such that
\par
\vspace*{1\baselineskip} 
\begin{equation}
    arg \, {\rm min} \, L(\theta) = \sum_{i=1}^N \frac{|D_i|} {|D|} L_i(\theta)
\end{equation}
\par
\vspace*{1\baselineskip} 
where $N$ is the total number of clients, $|D_i|$ is the number of samples on each client $i$, $|D|$ is the number of samples on all clients, $L(\theta)$ is the empirical loss on the global model, and $L_i(\theta)$ is the local empirical loss on each client $i$.
\par
\vspace*{1\baselineskip} 
In addition the FedAvg optimisation algorithm, other variants have been developed, such as FedBoost\cite{hamer2020fedboost}, FedNova\cite{wang2020tackling}, FedProx\cite{li2020federated} and FetchSGD\cite{rothchild2020fetchsgd}.

\subsubsection{Categorisation}
\begin{figure}
\centering
\includegraphics[width = 1\hsize]{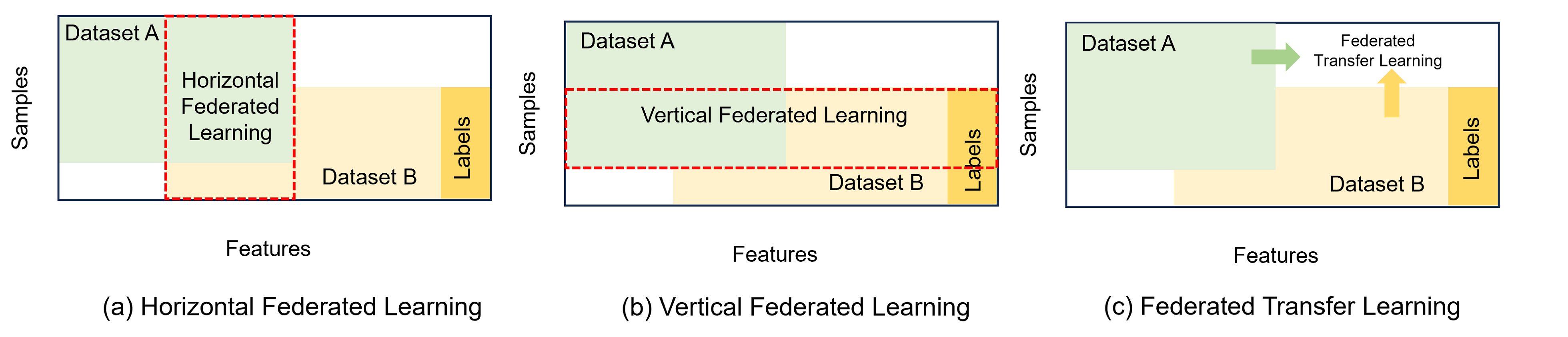}
\caption[Three main categories of federated learning: (a) Horizontal Federated Learning, (b) Vertical Federated Learning and (c) Federated Transfer Learning]{Three main categories of federated learning: (a) Horizontal Federated Learning, (b) Vertical Federated Learning and (c) Federated Transfer Learning.}
\label{fig:FL_Categories}
\end{figure}

Federated learning can be classified into three main categories, Horizontal Federated Learning, Vertical Federated Learning and Federated Transfer Learning, based on the distribution characteristics of the data in the feature and sample space\cite{yang2019federated}, as shown in Figure~\ref{fig:FL_Categories}.

\paragraph{Horizontal Federated Learning (HFL)} HFL refers to the scenario where datasets share an overlapping feature space but differ in sample space. The first use case of this type of learning is Google keyboard, where any participating users update the training parameters with the same features and transfer these parameters to the central cloud for model updates\cite{mcmahan2016federated}.

\paragraph{Vertical Federated Learning (VFL)} VFL applies to the scenario where datasets share an overlapping sample space but differ in feature space. For example, WeBank proposed a series of vertical federated learning models for object detection and anti-money laundering\cite{li2020review}\cite{liu2021fate}.

\paragraph{Federated Transfer Learning (FTL)} FTL is applicable for the scenarios where data distributed in different silos do not have many overlapping features and samples. This technique is an important extension of federated learning, since it does not place any restrictions on data distribution and allows for the knowledge gained from a specific domain to be applied to another related domain\cite{DBLP:journals/corr/abs-2010-15561}\cite{banabilah2022federated}.

\subsubsection{Limitations}
While Federated learning has offered benefits in terms of privacy protection, it also has certain limitations that need to be considered.
\paragraph{Data heterogeneity}
An imbalanced and non-IID dataset in federated learning can lead to biased global models and slower convergence rates, impacting overall model performance. Researchers have been actively working on modifying the existing aggregation methods or proposing new algorithms to solve these issues\cite{wang2020tackling}\cite{li2019convergence}.

\paragraph{System heterogeneity}

Federated networks involve clients with diverse hardware, network connectivity (4G, 5G and WiFi), and battery levels. \cite{li2020federated}. Variability in these factors can cause some clients to be unavailable during model updates.  Therefore, federated learning algorithms must be designed to be robust and adaptive to handle heterogeneous hardware effectively.

\paragraph{Communication overheads}

Federated learning also encounters communication challenges, with the need to transmit local model updates to the server and broadcast global model copies to all clients, leading to potential collisions and accuracy-cost tradeoffs. Researchers have proposed several approaches to minimise communication demands, by either data compression\cite{konevcny2016federated}\cite{kairouz2021advances} or sending only relevant information to the central server\cite{hsieh2017gaia}\cite{luping2019cmfl}.

\label{sec:dataimbalance}

\subsection{Federated Learning on Credit Risk Forecasting}
Federated learning is a relatively novel machine learning technique. This framework has been used in a variety of applications, such as Internet of Things (IoT)\cite{imteaj2019distributed}\cite{kwon2020multiagent}, blockchain\cite{pokhrel2020federated}, healthcare\cite{blanquer2020federated}\cite{elayan2021deep} and financial services\cite{yang2019ffd}. However, its application to credit risk forecasting has been relatively limited and underexplored. Kawa et al.\cite{kawa2019credit} proposed the use of federated learning framework for credit risk assessment but did not develop a real-life implementation based on this concept.
WeBank\cite{FedAIRisk}, a Internet bank of China, introduced logistic regression-based federated learning for the credit risk management of enterprises. Zheng et al.\cite{zheng2020vertical} applied a gradient-based federated learning framework with bounded constrains to perform credit risk forecasting. This framework achieved a prediction model with improved performance and reduced parameter-tuning process. More recently, Efe\cite{efe2021vertical} and He et al.\cite{he2023privacy} proposed federated learning-based approaches to collaboratively train a shared prediction model for credit risk assessment. These federated learning studies for credit risk assessment are compared in Table~\ref{tab:workcomparison}. It can be found that existing research predominantly revolves around logistic regression in federated learning, highlighting the need for exploring additional machine learning algorithms within this framework for credit risk forecasting.

\subsection{Federated Learning on Heterogeneous Data}
One of the primary challenges in federated learning is data heterogeneity (non-IID data). The most common non-IID data settings can be categorised into three types, quantity imbalance, class imbalance and feature imbalance\cite{kairouz2021advances}. Table~\ref{tab:compare_heterogeneous} summaries of existing federated learning approaches on non-IID data. The majority of prior research on simulating non-IID data primarily centers around class imbalance and feature imbalance. Many studies proposed various averaging methods to address the performance degradation caused by these imbalances\cite{yan2023label}\cite{khan2023precision}\cite{zhou2022fedfa}. There have been relatively few investigations into quantity imbalance. One example in this area is the work by Chung et al., who proposed FedISM to address quantity imbalance issue and improve the performance degradation caused by quantity skew\cite{chung2023fedism}. Certain studies have attempted to assess the impact of quantity imbalance on model performance\cite{liu2021adaptive}\cite{feki2021federated}. However, these studies have typically limited their analysis to comparing federated models in non-IID scenarios with either centralised models or federated models in IID scenarios. Additionally, these investigations have often only experimented with one or two imbalanced scenarios. Our work bridges this research gap, standing out as the pioneering effort in conducting comprehensive experiments involving three distinct model architectures and three datasets. Its primary aim is to explore the impact of quantity imbalances in federated learning. This is achieved through comparisons among federated, local, and centralised models across varying client numbers and data distributions.

\begin{table}
\caption{Comparison of Federated Learning studies on heterogeneous data.}
\label{tab:compare_heterogeneous}
\centering
\resizebox{\textwidth}{!}{
{\fontsize{8}{9}\selectfont
\begin{tabular}{p{0.4cm}p{0.4cm}p{1cm}p{1cm}
p{1.4cm}p{2.4cm}p{3.5cm}p{3.5cm}}
\toprule
 Study &
  Year &
  \begin{tabular}[c]{@{}l@{}}Number of \\ Datasets\end{tabular} &
  \begin{tabular}[c]{@{}l@{}}Number of\\ Clients\end{tabular} &
  Model Type &
  Data Imbalance Type &
  Simulated Cases &
  Conclusions \\ \midrule
\cite{liu2021adaptive} &
  2021 &
  2 &
  10 &
  Logistic regression and CNN &
  Quantity Imbalance &
  Simulated four cases (balance, weak quantity imbalance, strong quantity imbalance, class imbalance) among 10   workers &
  Training accuracy under imbalanced data distributions is close to that of the case with balanced data \\
\cite{feki2021federated} &
  2021 &
  1 &
  4 &
  CNN &
  Quantity Imbalance &
  Simulated two cases (one balanced and one imbalanced) among 4 workers &
  FL remains robust and shows comparable performance with a centralised learning despite imbalanced data distributions \\
\cite{qu2021experimental} &
  2021 &
  2 &
  4 &
  CNN &
  Quantity Imbalance Class Imbalance &
  Simulated four cases (one balanced and three imbalanced cases) among 4 workers &
  The proposed weighted average FedSGD can recover performance loss from quantity skew \\
\cite{zhou2022fedfa} &
  2022 &
  5 &
  100 &
  CNN &
  Feature Imbalance Class Imbalance&
  Simulated 10 clients with class distribution skew or feature distribution skew &
  The proposed FedFA outperforms baselines despite class imbalance and feature imbalance \\
\cite{khan2023precision} &
  2023 &
  1 &
  10 &
  ANN and CNN &
  Class Imbalance &
  Simulated 10 workers with class imbalanced local datasets &
  The suggested method resolves the challenge of class imbalance in non-IID data \\
\cite{chung2023fedism} &
  2023 &
  1 &
  10 &
  CNN &
  Quantity Imbalance &
  Simulated four cases (one balanced and three imbalanced cases) using Dirichlet distribution among 4 and 10 workers &
  The performance of FedAvg deteriorates as the data distributions become more imbalanced and the proposed FedISM can overcome this issue\\
\cite{yang2023fedrich} &
  2023 &
  3 &
  100 &
  CNN &
  Quantity Imbalance Class Imbalance &
  Simulated 10 workers with class imbalanced and quantity imbalanced datasets &
  The proposed FedRich mitigates the statistic heterogeneity and counterweights the performance degradation \\
\cite{yan2023label} &
  2023 &
  4 &
  5 &
  CNN and ViT-B &
  Class Imbalance &
  Simulated three cases (one balanced and two imbalanced cases) using the Dirichlet distribution among 5 workers &
  The proposed method outperforms existing federated self-supervised and semi-supervised FL methods  under non-IID and label deficient settings \\
\cite{casella2023experimenting} &
  2023 &
  2 &
  2, 4, 8 and 10 &
  CNN &
  Feature Imbalance Class Imbalance&
  Simulated four cases (one centralised, one balanced, one class imbalanced, one feature imbalanced cases) &
  Group and Layer Normalisation consistently outperform Batch Normalisation under IID and non-IID scenarios \\ \bottomrule
\end{tabular}
}
}
\end{table}

\subsection{Summary}
Our research addresses several key gaps in the current literature on federated learning overall, but specially within credit risk:

\begin{itemize}
  \item \textbf{Quantity Imbalance in Non-IID Data:} Existing studies have offered limited insights into data imbalance effects on federated learning model performance. Our research is the first to comprehensively investigate this aspect, comparing results with local and centralised baseline models for a comprehensive understanding of federated learning advantages."
  \item \textbf {Model Diversity:} A significant gap exists in terms of the diversity of models explored in credit risk forecasting under federated learning. The existing literature primarily focuses on logistic regression. In contrast, our work introduces three distinct model architectures (MLP, LSTM and XGBoost).
  \item \textbf{Dataset Diversity:} Previous credit risk forecasting studies in federated learning predominantly focused on binary classification datasets. Our work extends this by evaluating federated models across three datasets, comprising one multiclass and two binary datasets.
  
\end{itemize}

\section{Experiments}

\subsection{Experimental Design}
The investigation focuses on evaluating the performance of decentralised models under a Horizontal Federated Learning (HFL) setting where all the participating clients share the same set of features, considering different numbers of participating clients and data distributions. 

Three scenarios were simulated to compare federated learning models and non-federated learning models. Specifically, in the first scenario, a centralised model was used as a baseline under the assumption that a central server had access to the whole dataset. In the second scenario, a dataset was horizontally partitioned in a uniform and non-uniform way among the clients and each client trains its local model on local dataset. In the third scenario, a dataset was also randomly distributed among clients, assuming a collaborative approach to improve local models. 

The first two scenarios simulate non-federated learning baseline models, and the third simulates federated learning models. To analyse the effects of imbalanced data, we focused on a prominent client with 60\% or 80\% of the entire dataset, while the rest was distributed evenly among other clients. The detailed experimental design is listed in Table~\ref{tab:experimentalDesign}.
 
For the above three scenarios, MLP, LSTM and XGBoost models were utilised for Dataset 1 and Dataset 2. However, due to the absence of time series data in Dataset 3, only MLP and LSTM were employed since an LSTM model was not applicable to Dataset 3.

\begin{table}
\caption{Experimental design.}
\label{tab:experimentalDesign}
\centering
{\fontsize{7}{7}\selectfont
\begin{tabular}{@{}ccc@{}}
\toprule
                  & Number of Clients & Data Distribution (\%)                          \\ \midrule
Scenario 1 \\Centralised Model & 1             & 100                                      \\ \midrule
                  &              & 50-50                                   \\  
                  &  2             & 60-40                                                 \\
                  &               & 80-20                                                 \\ \cmidrule(l){2-3} 
                  &             & 34-33-33    \\
         &   3            & 60-20-20  \\
Scenario 2          &       & 80-10-10                                              \\ \cmidrule(l){2-3} 
Local Model      &              & 20-20-20-20-20                         \\
                     &  5             & 60-10-10-10-10                                        \\
                  &               & 80-5-5-5-5                                            \\ \cmidrule(l){2-3} 
                  &             & 10-10-10-10-10-10-10-10-10-10          \\
                  &    10           & 60-5-5-5-5-4-4-4-4-4                                  \\
                  &               & 80-3-3-2-2-2-2-2-2-2                                           \\ \midrule

                  &              & 50-50                                   \\
                  &  2             & 60-40                                                 \\
                  &               & 80-20                                                 \\ \cmidrule(l){2-3} 
                  &              & 34-33-33                                \\
                    & 3              & 60-20-20                                             \\
Scenario 3       &               & 80-10-10                                              \\ 
\cmidrule(l){2-3} 
Federated Model &  & 20-20-20-20-20\\
 
                &  5             & 60-10-10-10-10                                        \\
                  &               & 80-5-5-5-5                                            \\ \cmidrule(l){2-3} 
                  &             & 10-10-10-10-10-10-10-10-10-10            \\
                  &  10             & 60-5-5-5-5-4-4-4-4-4                                  \\
                  &               & 80-3-3-2-2-2-2-2-2-2                              
             \\ \bottomrule
\end{tabular}
}
\end{table}

In all scenarios, the dataset was randomly split into a training set containing 80\% of data samples and a test set containing 20\% of data samples. The training set was further split into $K$ = 2, 3, 5 and 10 subsets according to the designated data distributions. Hyperparameters were optimised using centralised models and subsequently applied to both local and federated models. 

\subsection{Evaluation}
For multiclass classification (Dataset 1), model performance was assessed using Mean Squared Error (MSE) and Area Under the Curve (AUC). The dataset extended the classification into investment-grade (BBB- and above) and non-investment-grade corporates, aligning with prior work\cite{hajek2014predicting}. This binary transformation allows the use of the AUC metric to evaluate model performance for Dataset 1.

For binary classification (Dataset 2 and Dataset 3), only the AUC metric was used for model evaluation.

Model performance underwent five-fold cross-validation. Local models were assessed for dominant (largest local dataset) and non-dominant clients (other small datasets). In balanced data distribution, metrics for both dominant and non-dominant clients were averaged from local datasets. In imbalanced data distribution, metrics for dominant clients came from the largest dataset, while for non-dominants, metrics were the average across the remainning datasets. All experiments were implemented using Flower\cite{Flower}.

\subsection{Federated Model Architecture}
In general, the models used in federated learning are modified from centralised models. Each client trains on its local dataset for a certain number of epochs before local models are aggregated centrally. This process repeats over a certain number of aggregation rounds until convergence. For the federated averaging strategy, FedAvg was adopted and 50\% of available clients were sampled during each round of aggregation, with a minimum of 2 participating clients. For XGBoost models on Dataset 1, 100\% of available clients were used because a 50\% sampling rate resulted in loss fluctuations after convergence.

\begin{figure}
\centering
\includegraphics[width = 0.9\hsize]{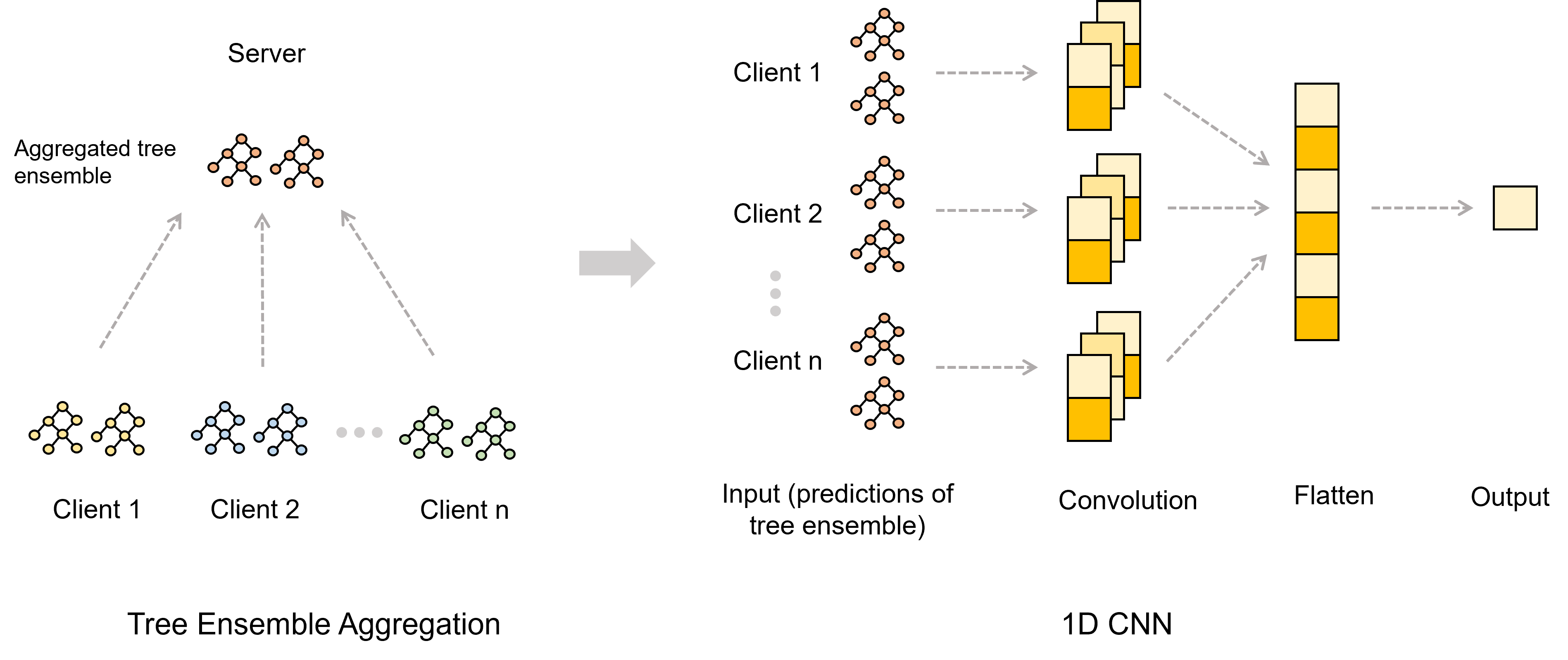}
\caption[Overview of XGBoost federated learning]{Overview of XGBoost federated learning.}
\label{fig:XGB+CNN}
\end{figure}

MLP and LSTM federated learning aligns with the model architecture outlined in Section \ref{model_architecture}, while XGBoost federated learning is slightly different. At the first aggregation round, each client constructs local XGBoost trees and send them to the central server. The server aggregates the trees and sends the aggregated trees to clients, which is similar to federated MLP and LSTM models. After the first round, each client uses the predictions of the aggregated tree ensembles as the input of a Convolutional Neural Network (CNN), which makes the learning rate of aggregated tree ensembles learnable. This method was first proposed by Ma et el. for binary classification and regression\cite{ma2023gradient}. In this research, we extended the model to accommodate multiclass classification, as both binary and multiclass classification problems were investigated. Figure~\ref{fig:XGB+CNN} shows an overview of XGBoost federated learning process.

\subsection{Datasets}

\subsubsection{Dataset 1}
The first dataset, Corporate Credit Rating Dataset from Wang and Ku\cite{wang2021utilizing}, contains 901 companies with 28 financial features from 2014 to 2016. We followed the S\&P credit rating scale, converting 22 distinct classes into {AAA, AA+, AA, AA-, A+, A, A-, BBB+, BBB, BBB-, BB+, BB, BB-, B+, B, B-, CCC+, CCC, CCC-, CC, C, D}. Ratings were further categorised into investment and non-investment groups for assessing default risk, framing it as a binary classification task.

The ANOVA F-test were performed to identify the most significant features, and the six with the lowest F-values were excluded before training.  Due to dataset imbalance, random upsampling was conducted to address class imbalances.

\subsubsection{Dataset 2}
The Home Credit Default Risk dataset from Kaggle addresses individual default risk\cite{home-credit-default-risk}, involving a binary classification task with target outputs of 0 or 1, indicating the individual repays the loan or not. With over 300,000 samples and 210 features, the dataset exhibits significant class imbalance, comprising 282,686 non-default records and 24,825 default records.

This dataset involves monthly time series data on previous loans. For the MLP and XGBoost models, a simple mean aggregation across all months was adopted for each applicant. Subsequently, the ANOVA F-test was performed to identify the most significant features.

\subsubsection{Dataset 3}
The Give Me Some Credit dataset from Kaggle focuses on individual default risk\cite{GiveMeSomeCredit}, comprising 150,000 data samples with 10 features. The dataset exhibits a substantial class imbalance, with 139,974 non-default records and only 10,026 default records.

For centralised baseline models, preprocessing filled missing values with the mean, followed by standardisation and upsampling before training. With only 10 features, no feature selection was done.

\section{Results}
\subsection{MLP Models}
\begin{figure}
\centering
\includegraphics[width = 0.66\hsize]{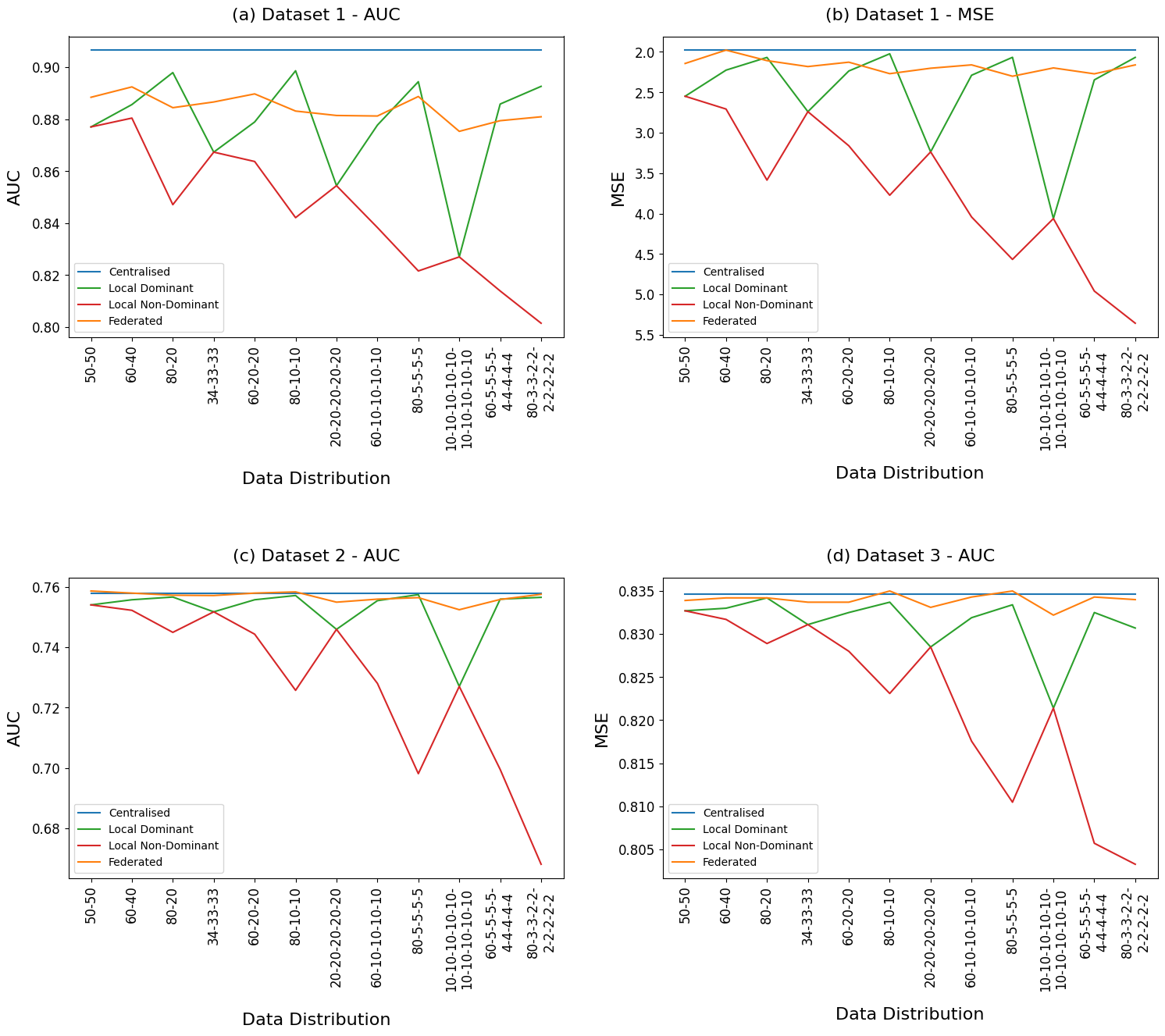}
\caption{MLP performance comparison.}
\label{fig:mlp_results}
\end{figure}

\begin{figure}
\centering
\includegraphics[width = 0.66\hsize]{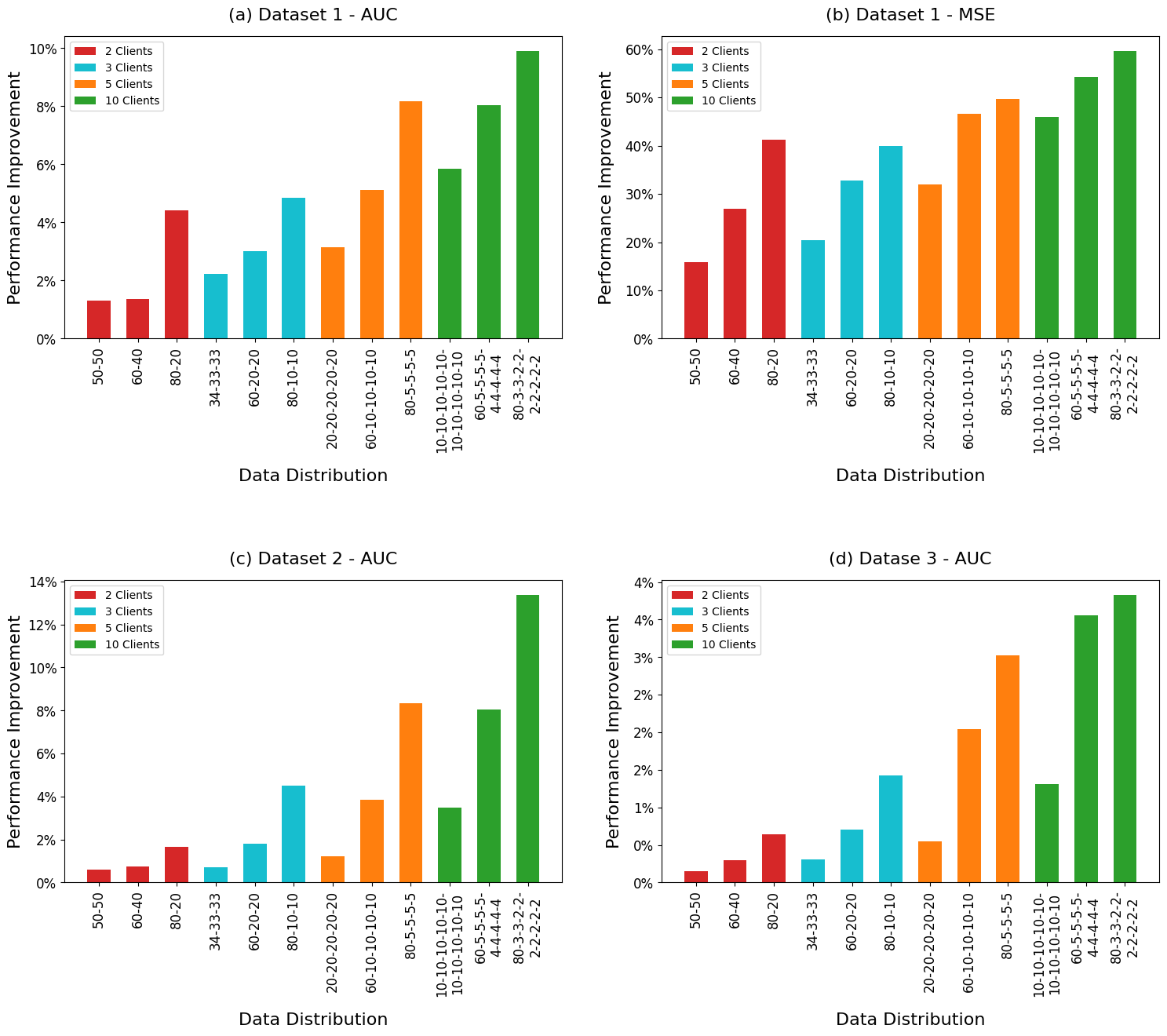}
\caption{Performance improvement of MLP with federated learning compared to local models on non-dominant clients.}
\label{fig:mlp_improve}
\end{figure}

The MLP performance across three datasets is summarised in Table~\ref{tab:MLP_results} and Figure~\ref{fig:mlp_results}. The centralised model demonstrated the highest performance across all experiments on Dataset 1, while federated models exhibited optimal performance on Dataset 2 and Dataset 3. Overall, centralised (blue lines in Figure~\ref{fig:mlp_results}) and federated models (yellow lines in Figure~\ref{fig:mlp_results}) exhibited comparable performance. It is also clear that while federated models were not necessarily better than local models on dominant clients  (green lines in Figure~\ref{fig:mlp_results}), all
of them outperformed their local non-dominant counterparts  (red lines in Figure~\ref{fig:mlp_results}). This advantage became more
pronounced with more clients and imbalanced data distribution, as shown in Figure ~\ref{fig:mlp_improve}. This trend is because federated learning maintained
stable performance, while local models on non-dominant clients deteriorated rapidly.

\begin{table}
\caption{MLP performance summary. Best performance across all distributions is underlined and best performance for the same distribution is highlighted in bold.}
\label{tab:MLP_results}
\centering
{\fontsize{7}{7}\selectfont
\begin{tabular}{ccccccccc}
\toprule
\multicolumn{3}{l}{}             & \multicolumn{2}{c}{Federated Learning} & \multicolumn{2}{c}{Local Dominant} & \multicolumn{2}{c}{Local Non-Dominant} \\ \midrule
Dataset &
Number of Clients & Data Distribution    & MSE             & AUC                   & MSE             & AUC             & MSE    & AUC    \\ \midrule
& 1             & 100 (Centralised)    & \underline{1.9765} & \underline{0.9066} &                 &                 &        &        \\ \cmidrule(l){2-9}
&              & 50-50                & \textbf{2.1435} & \textbf{0.8884}       & 2.5490          & 0.8770          & 2.5490 & 0.8770 \\
& 2             & 60-40                & \textbf{1.9783} & \textbf{0.8924}       & 2.2260          & 0.8856          & 2.7074 & 0.8804 \\
&              & 80-20                & 2.1084          & 0.8844                & \textbf{2.0689} & \textbf{0.8979} & 3.5865 & 0.8471 \\ \cmidrule(l){2-9}
&             & 34-33-33             & \textbf{2.1817} & \textbf{0.8866}      & 2.7398          & 0.8673          & 2.7398 & 0.8673 \\
Dataset 1 &3              & 60-20-20             & \textbf{2.1277} & \textbf{0.8897}      & 2.2366          & 0.8789          & 3.1617 & 0.8637 \\
&              & 80-10-10             & 2.2697          & 0.8831                & \textbf{2.0238} & \textbf{0.8986} & 3.7736 & 0.8421 \\ \cmidrule(l){2-9}
&              & 20-20-20-20-20       & \textbf{2.2024} & \textbf{0.8814}       & 3.2373          & 0.8544          & 3.2373 & 0.8544 \\
&5             & 60-10-10-10-10       & \textbf{2.1610} & \textbf{0.8812}       & 2.2894          & 0.8777          & 4.0424 & 0.8383 \\
&              & 80-5-5-5-5           & 2.3014          & 0.8887                & \textbf{2.0674} & \textbf{0.8944} & 4.5676 & 0.8216 \\ \cmidrule(l){2-9}
& & 10-10-10-10-10-10-10-10-10-10 & \textbf{2.1986}    & \textbf{0.8753}   & 4.0626           & 0.8270          & 4.0626             & 0.8270            \\
&10            & 60-5-5-5-5-4-4-4-4-4 & \textbf{2.2716} & 0.8794                & 2.3467          & \textbf{0.8858} & 4.9593 & 0.8139 \\
&              & 80-3-3-2-2-2-2-2-2-2 & 2.1614          & 0.8809                & \textbf{2.0685} & \textbf{0.8926} & 5.3570 & 0.8015 \\ \midrule \midrule

&1             & 100 (Centralised)             && 0.7577                &                 &                    \\ \cmidrule(l){2-9}
&              & 50-50                         && \underline {\textbf{0.7586}} && 0.7540          && 0.7540             \\
&2             & 60-40                         & &\textbf{0.7579}       && 0.7557          && 0.7522             \\
&              & 80-20                         && \textbf{0.7572}       && 0.7566          && 0.7449             \\ \cmidrule(l){2-9}
&              & 34-33-33                      && \textbf{0.7571}       && 0.7517          && 0.7517             \\
Dataset 2 &3             & 60-20-20                      & &\textbf{0.7579}       && 0.7557          && 0.7443             \\
&              & 80-10-10                      && \textbf{0.7583}       && 0.7571          && 0.7257             \\ \cmidrule(l){2-9}
 &             & 20-20-20-20-20                & &\textbf{0.7549}       && 0.7459          && 0.7459             \\
&5             & 60-10-10-10-10                & &\textbf{0.7559}       && 0.7554          && 0.7280             \\
&              & 80-5-5-5-5                    && 0.7564                && \textbf{0.7574} && 0.6981             \\ \cmidrule(l){2-9}
&              & 10-10-10-10-10-10-10-10-10-10 & &\textbf{0.7524}       && 0.7270          && 0.7270             \\
&10            & 60-5-5-5-5-4-4-4-4-4          && 0.7558                && \textbf{0.7559} && 0.6995             \\
&              & 80-3-3-2-2-2-2-2-2-2          && \textbf{0.7575}       && 0.7565          && 0.6681             
\\  \midrule \midrule

&1             & 100 (Centralised)    && 0.8346                &        &        \\ \cmidrule(l){2-9}
&              & 50-50                && \textbf{0.8339}       && 0.8327 && 0.8327 \\
&2             & 60-40                && \textbf{0.8342}       & &0.8330 && 0.8317 \\
&              & 80-20                && \textbf{0.8342}       & &0.8342 && 0.8289 \\ \cmidrule(l){2-9}
&              & 34-33-33             && \textbf{0.8337}       & &0.8311 && 0.8311 \\
Dataset 3 &3             & 60-20-20             && \textbf{0.8337}       && 0.8325 && 0.8280 \\
&              & 80-10-10             && \underline {\textbf{0.8350} }      && 0.8337 && 0.8231 \\ \cmidrule(l){2-9}
&              & 20-20-20-20-20       && \textbf{0.8331}       & &0.8285 && 0.8285 \\
&5             & 60-10-10-10-10       && \textbf{0.8343}       & &0.8319 && 0.8176 \\
&              & 80-5-5-5-5           && \underline{\textbf{0.8350}} && 0.8334 && 0.8105 \\ \cmidrule(l){2-9}
 &                    & 10-10-10-10-10-10-10-10-10-10 && \textbf{0.8322}    && 0.8214         && 0.8214             \\
&10            & 60-5-5-5-5-4-4-4-4-4 && \textbf{0.8343}       && 0.8325 && 0.8057 \\
 &             & 80-3-3-2-2-2-2-2-2-2 && \textbf{0.8340}       && 0.8307 && 0.8033 \\ 

\bottomrule
\end{tabular}
}
\end{table}

\subsection{LSTM Models}
\begin{figure}
\centering
\includegraphics[width = 0.86\hsize]{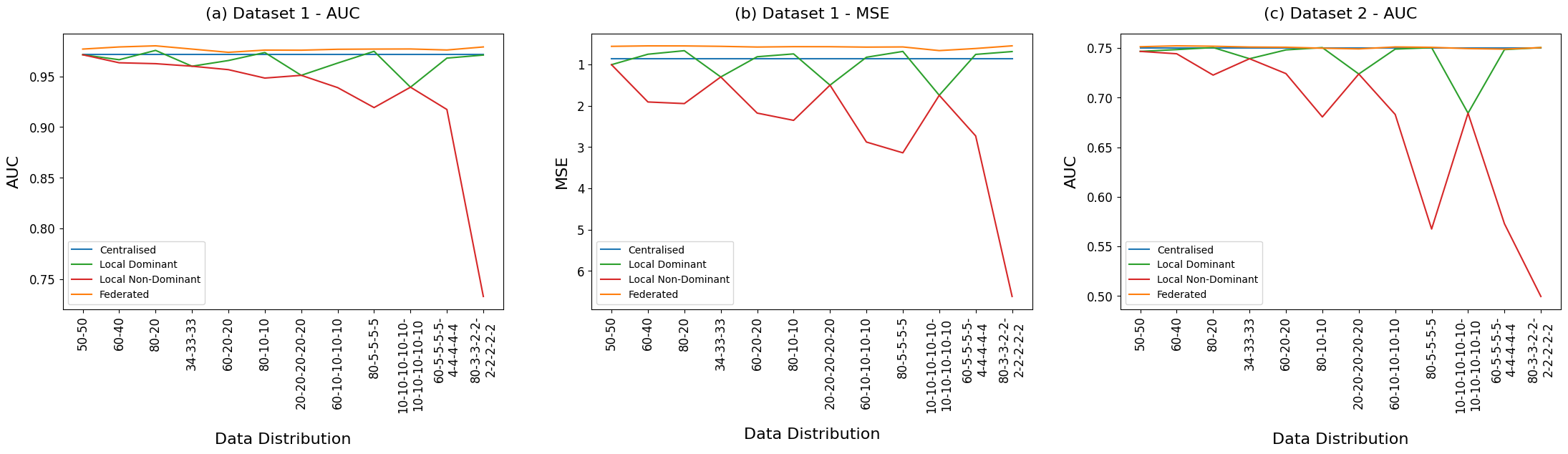}
\caption{LSTM performance comparison.}
\label{fig:lstm_results}
\end{figure}

\begin{figure}
\centering
\includegraphics[width = 0.86\hsize]{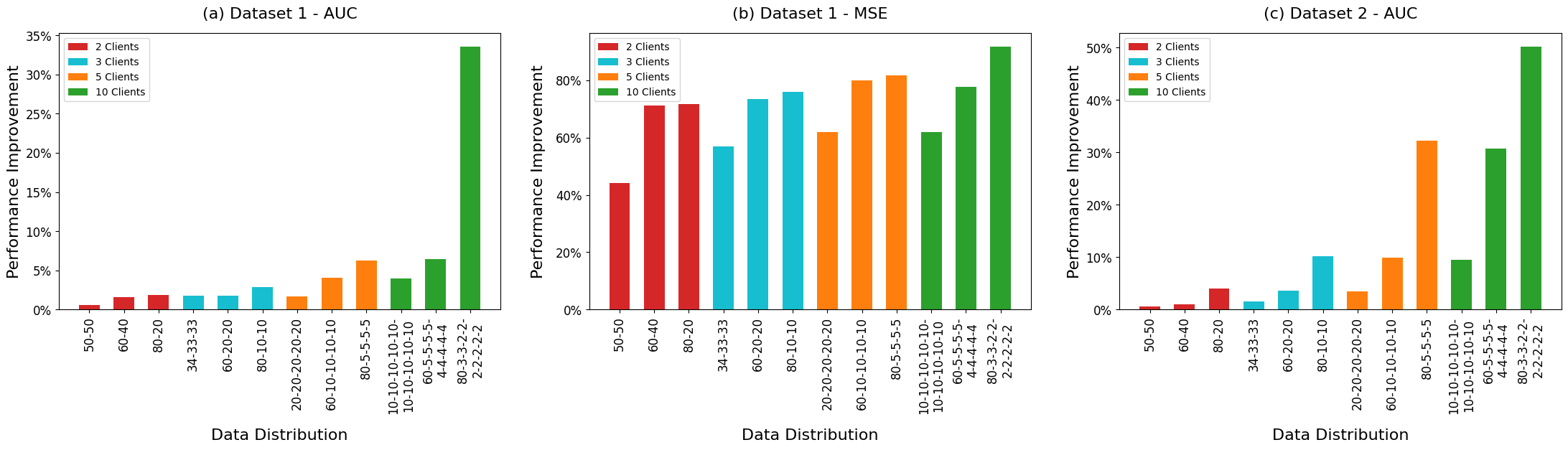}
\caption{Performance improvement of LSTM with federated learning compared to local models on non-dominant clients.}
\label{fig:lstm_improve}
\end{figure}

The LSTM performance of Dataset 1 and Dataset 2 is summarised in Table~\ref{tab:LSTM_results} and Figure~\ref{fig:lstm_results}. The optimal performance in all experiments were achieved through federated models, for both Dataset 1 and Dataset 2. These federated models generally outperformed local models in the case of dominant clients, with the exception of the 80-10-10 data distribution scenario in Dataset 2. Federated learning notably improved local model performance on non-dominant clients, especially with an increasing number of clients and imbalanced data distribution, as shown in Figure~\ref{fig:lstm_improve}.

\begin{table}
\caption{LSTM performance summary. Best performance across all distributions is underlined and best performance for the same distribution is highlighted in bold.}
\label{tab:LSTM_results}
\centering
{\fontsize{7}{7}\selectfont
\begin{tabular}{@{}ccccccccc@{}}
\toprule
\multicolumn{3}{l}{}             & \multicolumn{2}{c}{Federated Learning} & \multicolumn{2}{c}{Local Dominant} & \multicolumn{2}{c}{Local Non-Dominant} \\ \midrule
Dataset & Number of Clients & Data Distribution    & MSE                   & AUC                   & MSE    & AUC    & MSE    & AUC    \\ \midrule
&1             & 100 (Centralised)    & 0.8609                & 0.9714               &        &        &        &        \\ \cmidrule(l){2-9}
&              & 50-50                & \textbf{0.5617}       & \textbf{0.9769}       & 1.0064 & 0.9713 & 1.0064 & 0.9713 \\
 &2             & 60-40                & \underline{\textbf{0.5506}} & \textbf{0.9790}       & 0.7527 & 0.9664 & 1.9092 & 0.9634 \\
&              & 80-20                & \textbf{0.5517}       & \underline{\textbf{0.9801}} & 0.6672 & 0.9756 & 1.9480 & 0.9625 \\ \cmidrule(l){2-9}
&              & 34-33-33             & \textbf{0.5617}       & \textbf{0.9769}       & 1.3035 & 0.9600 & 1.3035 & 0.9600 \\
Dataset 1&3             & 60-20-20             & \textbf{0.5784}       & \textbf{0.9737}       & 0.8137 & 0.9656 & 2.1785 & 0.9566 \\
&              & 80-10-10             & \textbf{0.5695}       & \textbf{0.9759}       & 0.7450 & 0.9735 & 2.3546 & 0.9483 \\ \cmidrule(l){2-9}
&              & 20-20-20-20-20       & \textbf{0.5707}       & \textbf{0.9758}       & 1.5019 & 0.9597 & 1.5019 & 0.9597 \\
&5             & 60-10-10-10-10       & \textbf{0.5805}       & \textbf{0.9767}       & 0.8259 & 0.9631 & 2.8796 & 0.9389 \\
&              & 80-5-5-5-5           & \textbf{0.5762}       & \textbf{0.9769}       & 0.6848 & 0.9747 & 3.1410 & 0.9192 \\ \cmidrule(l){2-9}
& & 10-10-10-10-10-10-10-10-10-10 & \textbf{0.6648}    & \textbf{0.9770}   & 1.7470           & 0.9393          & 1.7470             & 0.9393            \\
&10            & 60-5-5-5-5-4-4-4-4-4 & \textbf{0.6139}       & \textbf{0.9760}       & 0.7568 & 0.9680 & 2.7321 & 0.9173 \\
&              & 80-3-3-2-2-2-2-2-2-2 & \textbf{0.5494}       & \textbf{0.9790}       & 0.6882 & 0.9710 & 6.6200 & 0.7329 \\ 
\midrule \midrule

&1             & 100 (Centralised)             && 0.7503                &                 &                    \\ \cmidrule(l){2-9}
&              & 50-50                         && \textbf{0.7513}       & &0.7466          && 0.7466             \\
&2             & 60-40                         && \underline{\textbf{0.7522}} && 0.7484          && 0.7442             \\
&              & 80-20                         && \textbf{0.7518}       & &0.7505          && 0.7227             \\ \cmidrule(l){2-9}
&              & 34-33-33                      && \textbf{0.7510}       & &0.7392          && 0.7392             \\
Dataset 2 &3             & 60-20-20              &        & \textbf{0.7508}     &  & 0.7481          && 0.7242             \\
&              & 80-10-10                      && 0.7498                & &\textbf{0.7504} && 0.6805             \\ \cmidrule(l){2-9}
&              & 20-20-20-20-20                && \textbf{0.7491}       & &0.7238          && 0.7238             \\
&5             & 60-10-10-10-10                && \textbf{0.7511}       & &0.7489          && 0.6830             \\
&              & 80-5-5-5-5                    && \textbf{0.7507}       & &0.7504          && 0.5675             \\ \cmidrule(l){2-9}
&              & 10-10-10-10-10-10-10-10-10-10 & &\textbf{0.7497}       & &0.6843          && 0.6843             \\
&10            & 60-5-5-5-5-4-4-4-4-4          & &\textbf{0.7489}       & &0.7484          && 0.5728             \\
&              & 80-3-3-2-2-2-2-2-2-2          & &\textbf{0.7504}       & &0.7504          && 0.4995             \\ 

\bottomrule
\end{tabular}
}
\end{table}

\subsection{XGBoost Models}
\begin{figure}
\centering
\includegraphics[width = 0.66\hsize]{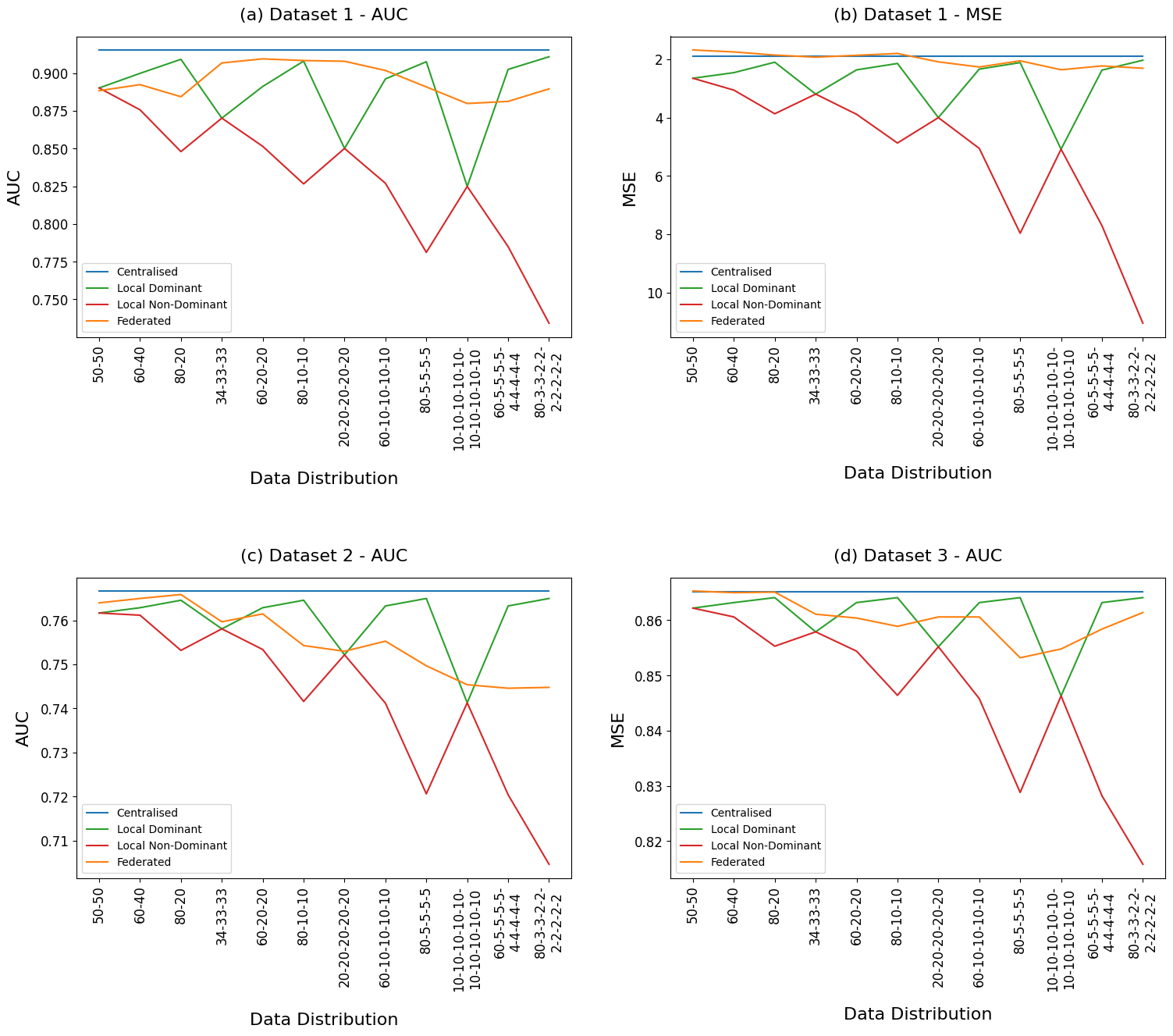}
\caption{XGBoost performance comparison.}
\label{fig:xgboost_results}

\end{figure}

\begin{figure}
\centering
\includegraphics[width = 0.66\hsize]{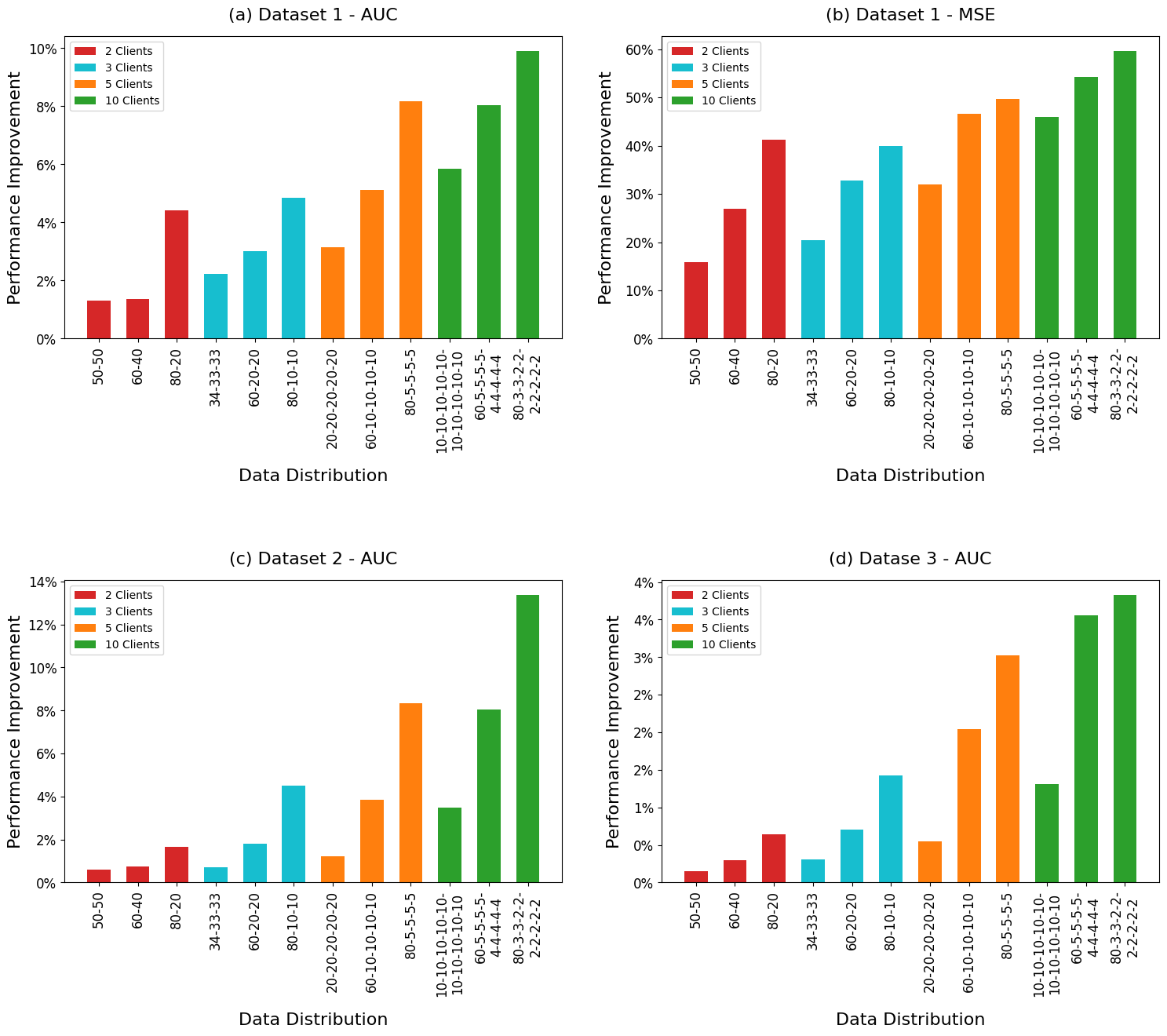}
\caption{Performance improvement of XGBoost with federated learning compared to local models on non-dominant clients.}
\label{fig:xgboost_improve}
\end{figure}

The XGBoost  performance across three datasets is summarised in Table~\ref{tab:XGBoost_results} and Figure~\ref{fig:xgboost_results}.
The optimal performance varied across datasets, with the centralised model demonstrating superiority for Dataset 2, while federated models excelled for Dataset 1 and Dataset 3. Federated models (yellow lines in Figure~\ref{fig:xgboost_results}) consistently outperformed local models on non-dominant clients (red lines in Figure~\ref{fig:xgboost_results}) . This
trend was particular evident with more clients and imbalanced data distributions, as shown in Figure~\ref{fig:xgboost_improve}. It is worth noting, however, that the performance of federated models did not always outperform that of local models for dominant clients (green lines in Figure~\ref{fig:xgboost_results}) across all three datasets.

\begin{table}
\caption{XGBoost performance summary. Best performance across all distributions is underlined and best performance for the same distribution is highlighted in bold.}
\label{tab:XGBoost_results}
\centering
{\fontsize{7}{7}\selectfont
\begin{tabular}{@{}ccccccccc@{}}
\toprule
 &                               & \multicolumn{2}{c}{Federated Learning} & \multicolumn{3}{c}{Local Dominant} & \multicolumn{2}{c}{Local Non-Dominant} \\ \midrule
Dataset & Number of Clients & Data Distribution    & MSE                   & AUC                   & MSE             & AUC             & MSE     & AUC    \\ \midrule
&1             & 100 (Centralised)    & 1.9109                & 0.9154                &                 &                 &         &        \\ \cmidrule(l){2-9}
&              & 50-50                & \underline{\textbf{1.6839}} & \textbf{0.9086}       & 2.6536          & 0.8902          & 2.6536  & 0.8902 \\
&2             & 60-40                & \textbf{1.7511}       & \underline{\textbf{0.9168}} & 2.4610          & 0.8999          & 3.0601  & 0.8757 \\
&              & 80-20                & \textbf{1.8595}       & \textbf{0.9102}       & 2.1033          & 0.9092          & 3.8706  & 0.8480 \\ \cmidrule(l){2-9}
&              & 34-33-33             & \textbf{1.9288}       & \textbf{0.9068}       & 3.1959          & 0.8702          & 3.1959  & 0.8702 \\
Dataset 1 &3             & 60-20-20             & \textbf{1.8678}       & \textbf{0.9095}       & 2.3660          & 0.8912          & 3.8887  & 0.8515 \\
&              & 80-10-10             & \textbf{1.8041}       & \textbf{0.9084}       & 2.1461          & 0.9080          & 4.8726  & 0.8266 \\ \cmidrule(l){2-9}
&              & 20-20-20-20-20       & \textbf{2.0915}       & \textbf{0.9079}       & 4.0029          & 0.8501          & 4.0029  & 0.8501 \\
&5             & 60-10-10-10-10       & \textbf{2.2671}       & \textbf{0.9018}       & 2.3401          & 0.8962          & 5.0567  & 0.8270 \\
&              & 80-5-5-5-5           & \textbf{2.0564}       & 0.8909                & 2.1170          & \textbf{0.9076} & 7.9649  & 0.7812 \\ \cmidrule(l){2-9}
& & 10-10-10-10-10-10-10-10-10-10 & \textbf{2.3623}    & \textbf{0.8799}   & 5.0851           & 0.8249          & 5.0851             & 0.8249            \\
&10            & 60-5-5-5-5-4-4-4-4-4 & \textbf{2.2304}       & 0.8813                & 2.3741          & \textbf{0.9025} & 7.7057  & 0.7850 \\
              && 80-3-3-2-2-2-2-2-2-2 & 2.3105                & 0.8896                & \textbf{2.0355} & \textbf{0.9109} & 11.0488 & 0.7342 \\ 
              \midrule \midrule

&1             & 100 (Centralised)             && \underline{0.7667}       &                 &                    \\ \cmidrule(l){2-9}
&              & 50-50                         && \textbf{0.7640}    && 0.7617          && 0.7617             \\
&2             & 60-40                         && \textbf{0.7650}    && 0.7629          && 0.7612             \\
&              & 80-20                         && \textbf{0.7659}    && 0.7646          && 0.7532             \\ \cmidrule(l){2-9}
&              & 34-33-33                      && \textbf{0.7597}    && 0.7581          && 0.7581             \\
Datase 2&3             & 60-20-20                      && 0.7615             & &\textbf{0.7629} && 0.7534             \\
&              & 80-10-10                      && 0.7543             & &\textbf{0.7646} && 0.7416             \\ \cmidrule(l){2-9}
&              & 20-20-20-20-20                && \textbf{0.7530}    && 0.7522          && 0.7522             \\
&5             & 60-10-10-10-10                && 0.7553             & &\textbf{0.7633} && 0.7412             \\
&              & 80-5-5-5-5                    & &0.7497             & &\textbf{0.7650} && 0.7206             \\ \cmidrule(l){2-9}
&              & 10-10-10-10-10-10-10-10-10-10 & &\textbf{0.7454}    && 0.7413          && 0.7413             \\
&10            & 60-5-5-5-5-4-4-4-4-4          & &0.7446             & &\textbf{0.7633} && 0.7204             \\
&              & 80-3-3-2-2-2-2-2-2-2          & &0.7448             & &\textbf{0.7650} && 0.7046             \\ 
\midrule \midrule

&1             & 100 (Centralised)    && 0.8652                &                 &        \\ \cmidrule(l){2-9}
 &             & 50-50                && \underline{\textbf{0.8653}} && 0.8622          && 0.8622 \\
&2             & 60-40                && \textbf{0.8650}       && 0.8632          && 0.8606 \\
&              & 80-20                && \textbf{0.8651}       && 0.8641          && 0.8553 \\ \cmidrule(l){2-9}
&              & 34-33-33             && \textbf{0.8611}      & & 0.8579          && 0.8579 \\
Dataset 3&3             & 60-20-20             && 0.8604                && \textbf{0.8632} && 0.8544 \\
&              & 80-10-10             && 0.8589                && \textbf{0.8641} && 0.8464 \\ \cmidrule(l){2-9}
&              & 20-20-20-20-20       && \textbf{0.8606}       && 0.8552          && 0.8552 \\
&5             & 60-10-10-10-10       && 0.8606                && \textbf{0.8632} && 0.8458 \\
&              & 80-5-5-5-5           && 0.8532                && \textbf{0.8641} && 0.8288 \\ \cmidrule(l){2-9}
&                     & 10-10-10-10-10-10-10-10-10-10 && \textbf{0.8548}      && 0.8463         && 0.8463               \\
&10            & 60-5-5-5-5-4-4-4-4-4 && 0.8584                && \textbf{0.8632} && 0.8282 \\
 &             & 80-3-3-2-2-2-2-2-2-2 && 0.8614                && \textbf{0.8641} && 0.8158 
\\
              \bottomrule
\end{tabular}
}
\end{table}

\section{Discussion}
In this section, our exclusive focus lies on the Area Under the Curve (AUC), which was employed for all datasets investigated in this research. Within this section, we thoroughly examine the performance of federated models in comparison to three types of baseline models: centralised models, local models on dominant clients, and local models on non-dominant clients. Our dedicated analysis aims to provide a comprehensive understanding of how data imbalance influences model performance under the framework of federated learning.

\subsection{Comparison with Local Models (Non-Dominants)}

\begin{figure}
\centering
\includegraphics[width = 0.6\hsize]{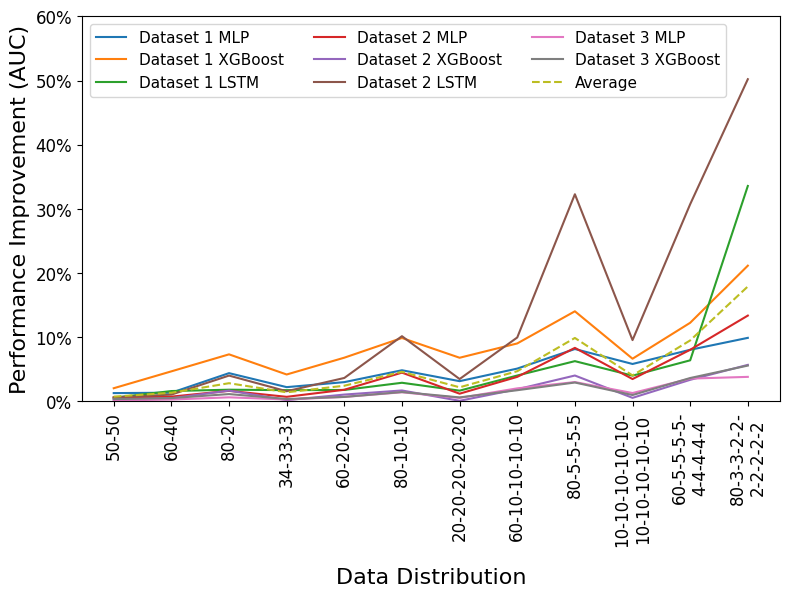}
\caption{Performance improvement summary of federated models compared to local models on non-dominant clients. Solid lines represent  the performance improvement for each model. The dashed line represents the average improvement across all models.}
\label{fig:non-dominant_compare}
\end{figure}

\begin{figure}
\centering
\includegraphics[width = 1\hsize]{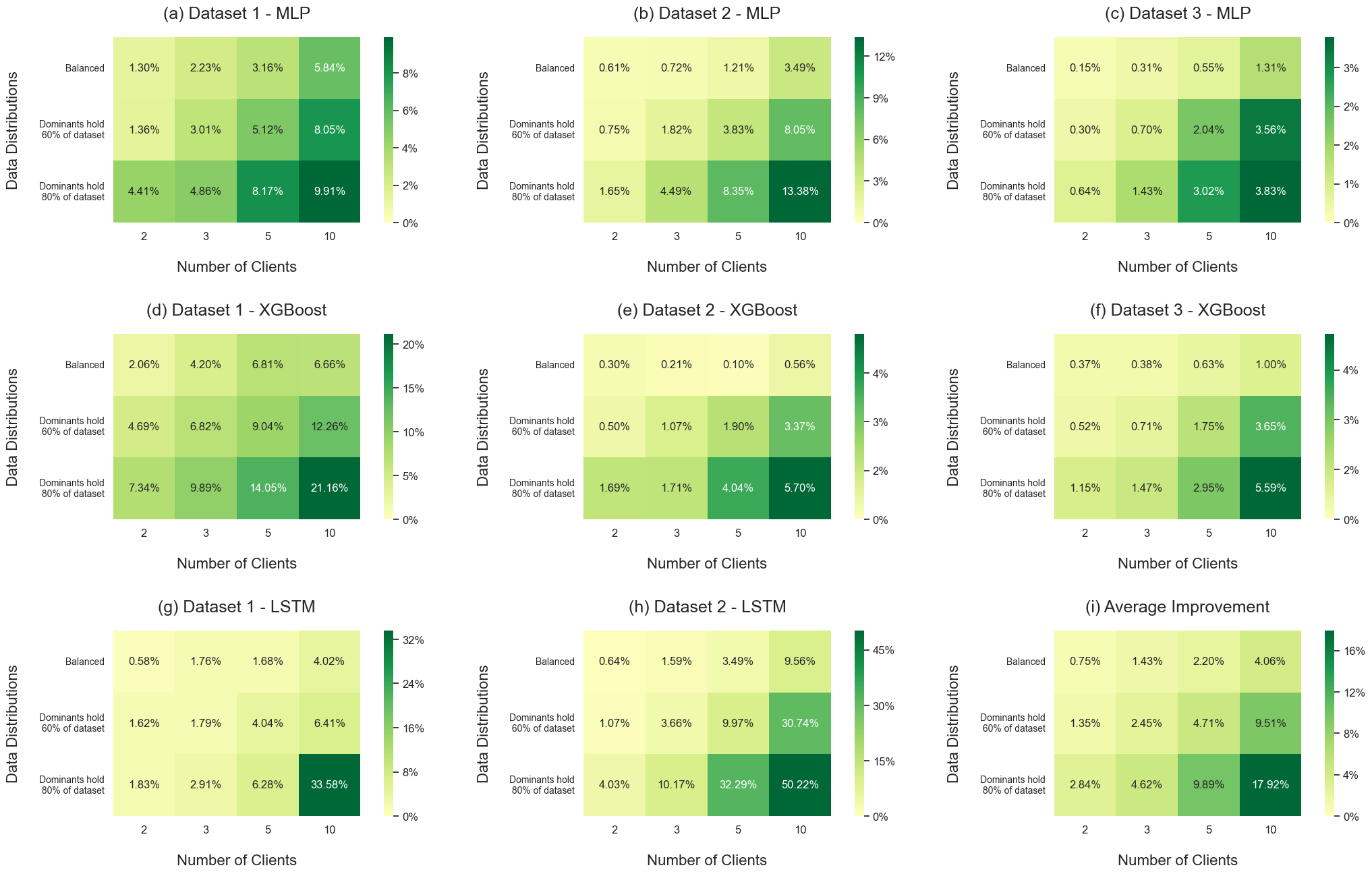}
\caption{Heatmap of performance improvement. Green denotes federated models are superior to local models on non-dominant clients.}
\label{fig:heatmap_nondominant}
\end{figure}

\begin{figure}
\centering
\includegraphics[width = 0.6\hsize]{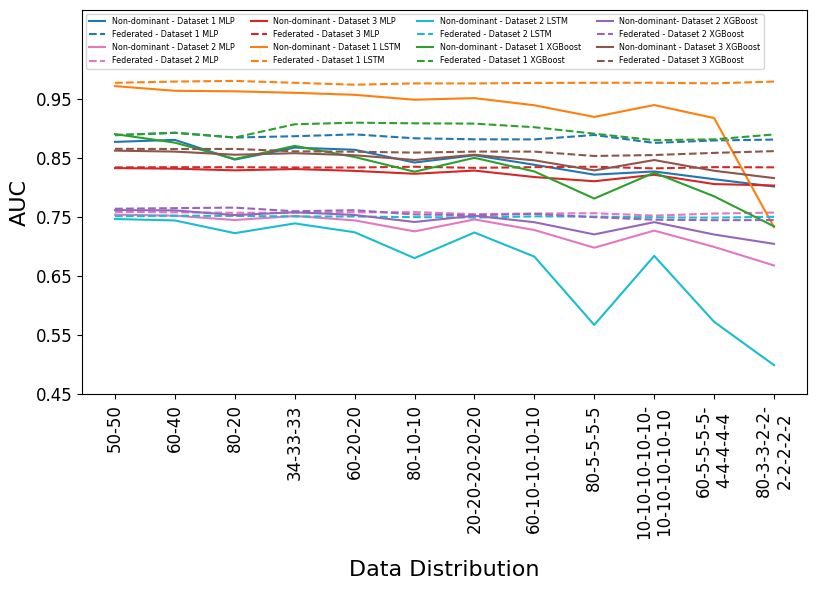}
\caption{Comparison of federated models and local models on non-dominant clients. Solid lines represent local models on non-dominant clients while dashed lines represent federated models.}
\label{fig:non-dominant_absolute_compare}
\end{figure}

The performance improvements for local models on non-dominant clients are shown in Figure~\ref{fig:non-dominant_compare} and Figure~\ref{fig:heatmap_nondominant}. It is evident that the model shows greater improvement as the number of clients increases and data distribution becomes more imbalanced. This is because the federated models (dashed lines in Figure~\ref{fig:non-dominant_absolute_compare}) are able to maintain their performance, whereas local models (solid lines in Figure~\ref{fig:non-dominant_absolute_compare}) degrade quickly under imbalanced conditions. Figure~\ref{fig:non-dominant_compare} shows that the most significant performance improvement is achieved when there are 10 clients, particularly in the case of the most imbalanced data distribution. In this scenario, one client possesses 80\% of the entire dataset, while the other nine clients hold local datasets that are evenly partitioned from the rest of the dataset. Under this condition, federated learning shows its highest potential for performance improvement. The achieved average gains are noteworthy, reflecting an average improvement of 17.92\% in the Area Under the Curve (AUC). These improvements have the potential to yield substantial cost savings in credit risk modeling.

\subsection{Comparison with Local Models (Dominants)}

\begin{figure}
\centering
\includegraphics[width = 0.6\hsize]{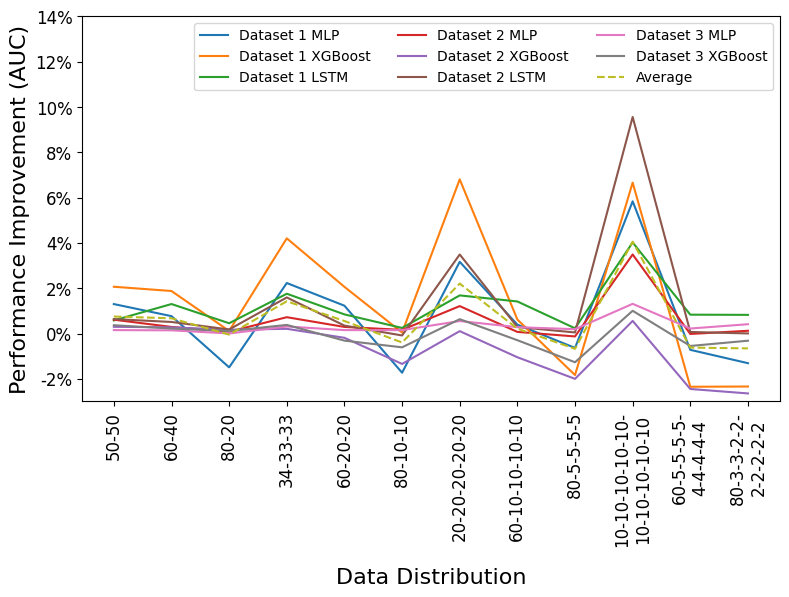}

\caption{Performance improvement summary of federated models compared to local models on dominant clients.}
\label{fig:dominant_compare}
\end{figure}

\begin{figure}
\centering
\includegraphics[width = 1\hsize]{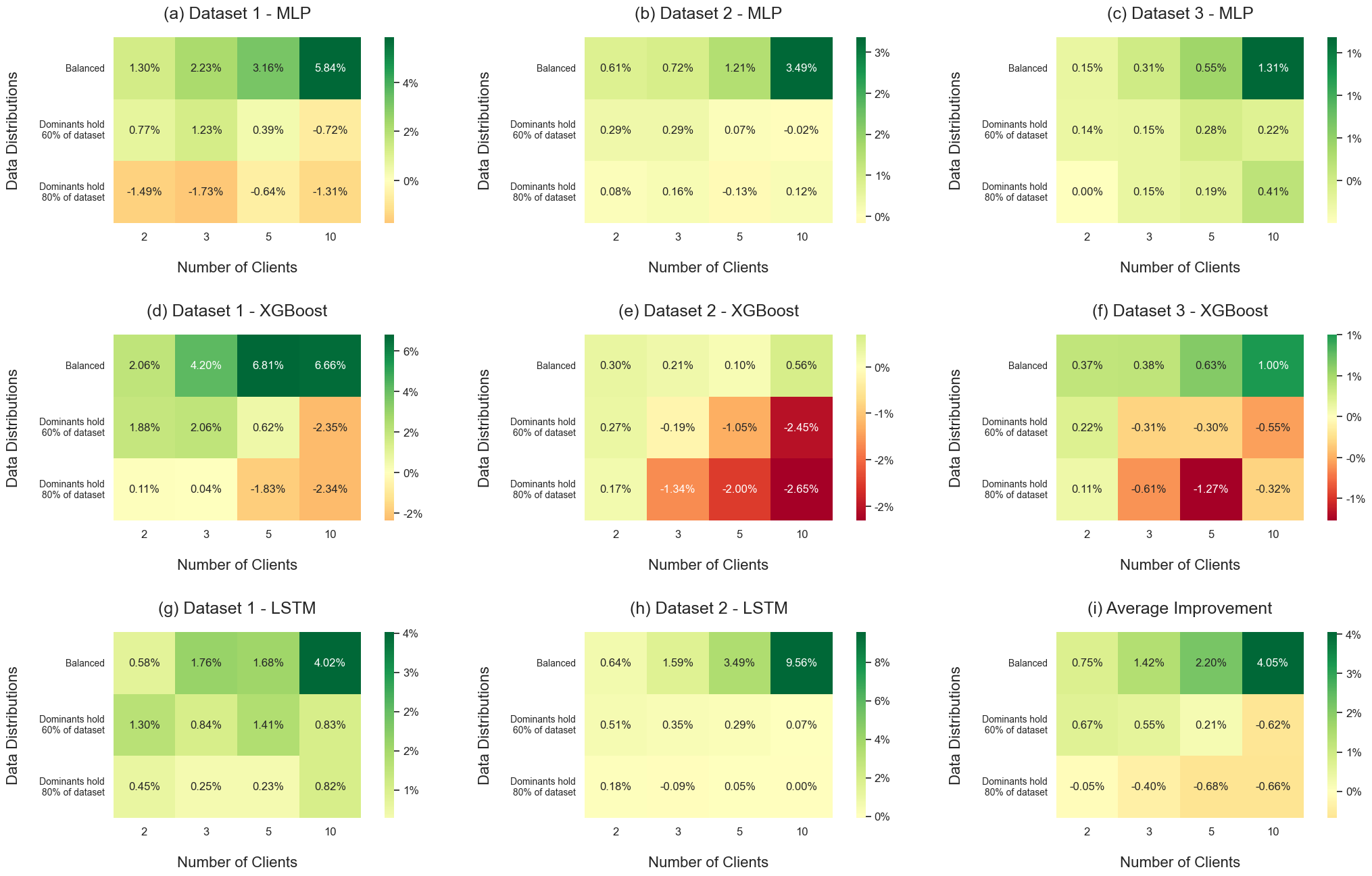}
\caption{Heatmap of performance improvement. Green denotes federated models are superior to local models on dominant clients, while red denotes the superiority of local models.}
\label{fig:heatmap_dominant}

\end{figure}

\begin{figure}
\centering
\includegraphics[width = 0.6\hsize]{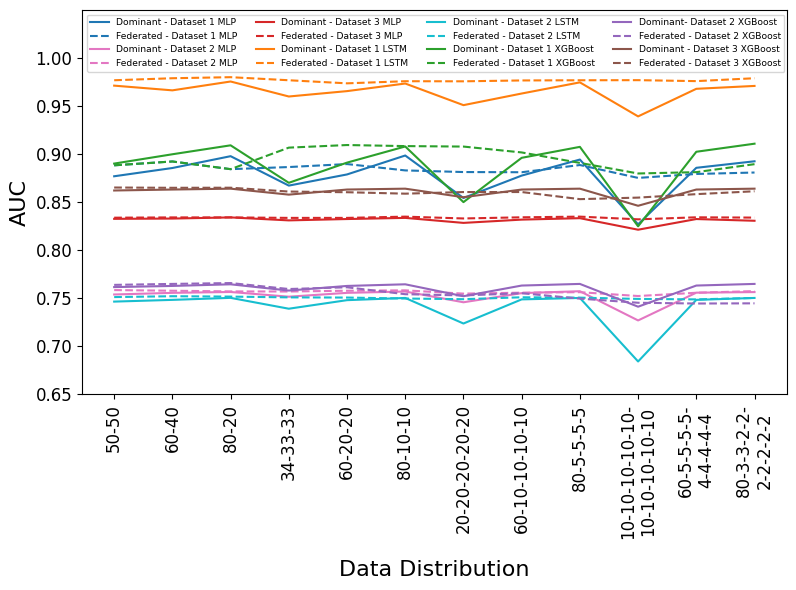}
\caption{Comparison of federated models and local models on dominant clients. Solid lines represent local models on dominant clients, while dashed lines represent federated models.}
\label{fig:dominant_absolute_compare}
\end{figure}

The performance improvements for local models on dominant clients are shown in Figure~\ref{fig:dominant_compare} and Figure~\ref{fig:heatmap_dominant}. Federated models are not consistently superior than local models, particularly in more imbalanced situations. This can be attributed to the fact that dominant clients possess a larger share of data in these scenarios, resulting in more representative models and better performance. In contrast, the performance of federated models (dashed lines in Figure~\ref{fig:dominant_absolute_compare}) remains relatively stable under different conditions. This indicates that clients with substantial larger datasets may lack a strong motivation to engage in federated learning. Therefore, these clients should be provided with special incentives, such as getting paid for their participation, while non dominants are expected to pay for it.

\subsection{Comparison with Centralised Models}
\begin{figure}
\centering
\includegraphics[width = 0.6\hsize]{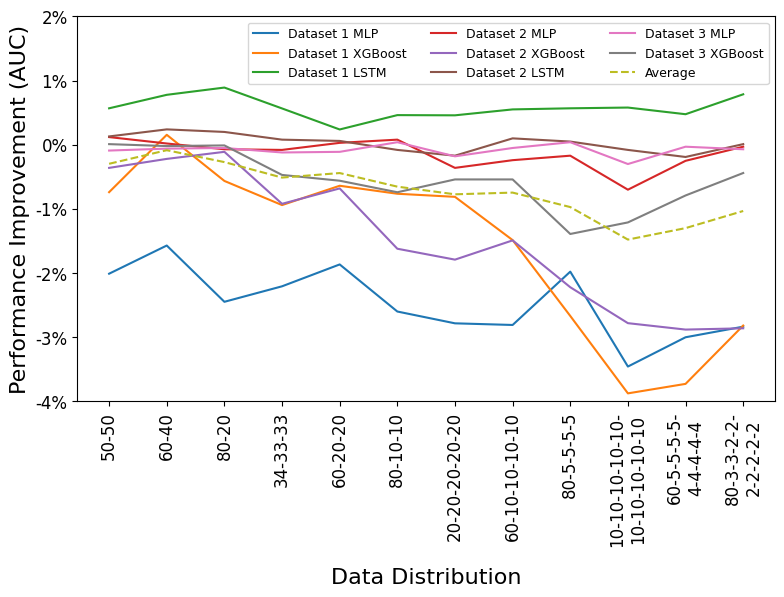}
\caption{Performance improvement summary of federated models compared to centralised models. Solid lines represent  the performance improvement for each model. The dashed line represents the average improvement across all models.}
\label{fig:central_compare}
\end{figure}

\begin{figure}
\centering
\includegraphics[width = 0.6\hsize]{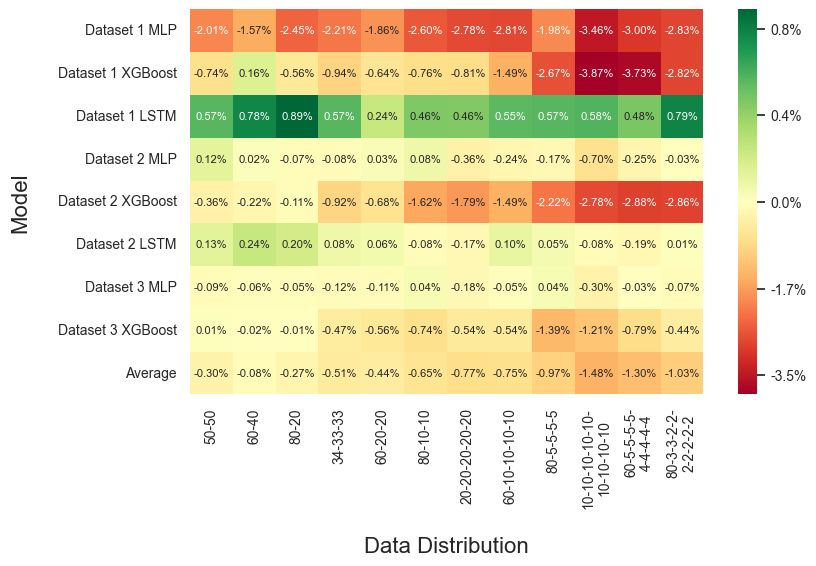}
\caption{Heatmap of performance improvement. Green denotes federated models are superior to centralised models, while red denotes the superiority of centralised models.}
\label{fig:heatmap_central}
\end{figure}

\begin{figure}
\centering
\includegraphics[width = 0.6\hsize]{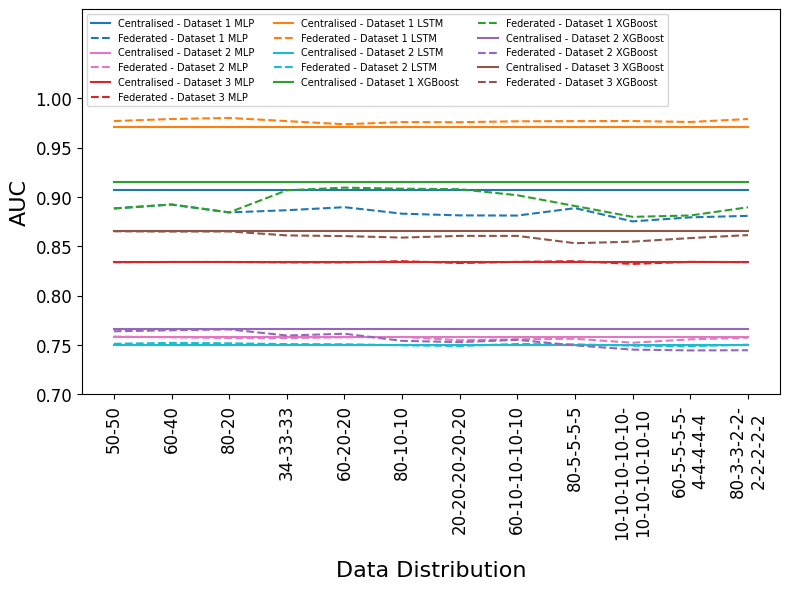}
\caption{Comparison of centralised and federated models. Solid lines represent centralised models, while dashed lines represent federated models.}
\label{fig:centralised_value}
\end{figure}

Figure~\ref{fig:central_compare} and Figure~\ref{fig:heatmap_central} provide a comparison between federated and centralised models, showing that there is no definitive guarantee of superiority for either federated or centralised models. They generally exhibit comparable performance, and the average performance gap (dashed line in Figure~\ref{fig:central_compare}) remains modest, with average differences below 1.5\%. It is worth noting that hyperparameters were fine-tuned for centralised models and then applied to federated models, which might explain why federated models may not perform as well as centralised ones in some scenarios. Additionally, Figure~\ref{fig:centralised_value} indicates that federated models (dashed lines) are robust since they do not exhibit significant performance degradation when the number of clients increases and data distributions become more imbalanced.

\section{Conclusion}
In this work, we have demonstrated the potential benefits of federated learning for use in credit risk forecasting through extensive experiments with three model architectures (two neural network-based techniques, MLP and LSTM, along with a tree ensemble technique, XGBoost) on three datasets. 

This research demonstrates that federated learning always outperform their local models in the case of non-dominant clients, while it may not consistently outperform the local models in scenarios where dominant clients possess substantial local datasets, especially when these clients have more data available. Furthermore, our study highlights the robustness and scalability of federated learning, even in the face of data imbalance. Federated learning achieves competitive performance compared to centralised models, underscoring its suitability for real-world applications.

These results hold substantial commercial implications. Although many people have been experimenting with federated learning, there are still very few commercial cases in which this has been delivered in practice. On average, for non dominant clients, federated learning could improve model performance by up to 17.92\%. This improvement could translate into substantial savings in credit risk modelling.

Additionally, this work demonstrates that when a consortium of organisations decides to cooperate, clients with much more data do not have a good incentive to participate since federated learning will not improve much over its existing results. This means that such type of clients should receive special incentives to participate in any cooperation. For example, potentially it should get paid for participating while non dominants should pay for it.

In conclusion, this study stands as a pioneering endeavor dedicated to investigating the impact of data imbalance on federated learning and delivers valuable commercial implications. We have demonstrated that federated learning offers a robust and scalable solution, particularly beneficial for non-dominant clients and in scenarios with data imbalance. These findings not only advance our understanding of federated learning but also offer actionable insights for its implementation in real-world scenarios.

Future work should involve fine-tuning hyperparameters specifically for federated models, rather than adopting those optimised for centralised models as done in this study. A comparison between federated and centralised models using tuned hyperparameters would provide valuable insights for further analysis. Additionally, future investigations could focus on the averaging algorithm to enhance both model performance and computation time, particularly when dealing with non-IID data.

\section*{Acknowledgments}
We thank Ovidiu Serban and Szymon Padlewski for their discussions and advice on the research.

\bibliographystyle{unsrt}  
\bibliography{main}

\end{document}